\documentclass[10pt,twocolumn,letterpaper]{article}

\usepackage[pagenumbers]{cvpr} 

\usepackage{graphicx}
\usepackage{amsmath}
\usepackage{amssymb}
\usepackage{booktabs}
\usepackage{enumitem}

\DeclareMathOperator*{\argmin}{arg\,min}
\makeatletter
\renewcommand\paragraph{\@startsection{paragraph}{4}{\z@}%
                                    {0.2ex \@plus1ex \@minus.1ex}%
                                    {-1em}%
                                    {\normalfont\normalsize\bfseries}}
\makeatother

\usepackage[pagebackref,breaklinks,colorlinks]{hyperref}

\usepackage[capitalize]{cleveref}
\crefname{section}{Sec.}{Secs.}
\Crefname{section}{Section}{Sections}
\Crefname{table}{Table}{Tables}
\crefname{table}{Tab.}{Tabs.}

\def\cvprPaperID{11260} 
\def\confName{CVPR}
\def\confYear{2022}

\begin{document}

\title{ Cross-modal Map Learning for Vision and Language Navigation }

\author{Georgios Georgakis, Karl Schmeckpeper, Karan Wanchoo, Soham Dan, \\ Eleni Miltsakaki, Dan Roth, Kostas Daniilidis\\
University of Pennsylvania\\
{\tt\small \{ggeorgak,karls,kwanchoo,sohamdan,elenimi,danroth,kostas\}@seas.upenn.edu}
\\
{\small Project webpage: \url{https://ggeorgak11.github.io/CM2-project/}}
\vspace{-2mm}
}
\maketitle

\begin{abstract}
We consider the problem of Vision-and-Language Navigation (VLN). The majority of current methods for VLN are trained end-to-end using either unstructured memory such as LSTM, or using cross-modal attention over the egocentric observations of the agent. In contrast to other works, our key insight is that the association between language and vision is stronger when it occurs in explicit spatial representations.  In this work, we propose a cross-modal map learning model for vision-and-language navigation that first learns to predict the top-down semantics on an egocentric map for both observed and unobserved regions, and then predicts a path towards the goal as a set of waypoints. In both cases, the prediction is informed by the language through cross-modal attention mechanisms. We experimentally test the basic hypothesis that language-driven navigation can be solved given a map, and then show competitive results on the full VLN-CE benchmark.
\vspace{-2.5mm}
\end{abstract}
\definecolor{ao}{rgb}{0.0, 0.5, 0.0}
\definecolor{darkorange}{rgb}{1.0, 0.55, 0.0}

\section{Introduction}

For mobile robots to be able to operate together with humans, they must be able to execute tasks that are defined not in the form of machine-readable scripts but rather in the form of human instructions. A very basic but challenging task is going from A to B. While robots have been quite successful in executing this task using metric representations, it has been more challenging for robots to execute semantic tasks like ``go to the kitchen sink'' or follow instructions that describe a path and associate actions with natural language, defined as the Vision-and-Language Navigation (VLN) task~\cite{anderson2018vision,krantz2020beyond,ku2020room}. In VLN, the robot is given instructions and has to reach a goal making use of images of the environment that it can acquire along the way.  

The dominant approach for VLN tasks has been using end-to-end pipelines from images and instructions to  actions~\cite{hao2020towards,fried2018speaker,krantz2021waypoint,krantz2020beyond}. While they can be attractive due to their simplicity, they are expected to implicitly learn end-to-end all navigation components such as mapping, planning, and control, and thus often require considerable amounts of training data.
This approach to designing navigation systems is in direct contrast to
research on human spatial navigation, which has shown that humans and other species build map-like representations of the environment to accomplish way-finding \cite{okeefe,tolman}. 
However, multiple findings have shown that the ability to build cognitive maps and acquire spatial knowledge deteriorates when humans exclusively use ready to drive or walk paths to a goal~\cite{brugger2019does}.
On the other hand, studies have shown that humans build better spatial representations when presented with landmark-based navigation instructions rather than full paths~\cite{wunderlich20biorxiv}. Such spatial representations enable the recall of landmarks on an egocentric map weeks after the experiment. While this does not prove that humans build a map during wayfinding when following semantic instructions, it is a strong indication that they can anchor landmarks and other semantics to a map that they easily recall. 
Research in learning of mapping and planning in computer vision and robotics~\cite{gupta2017cognitive} has also shown that an end-to-end system 
encompasses semantic maps that naturally emerge in the learning process.  
\begin{figure}[t]
    \centering
    \includegraphics[width=0.9\linewidth]{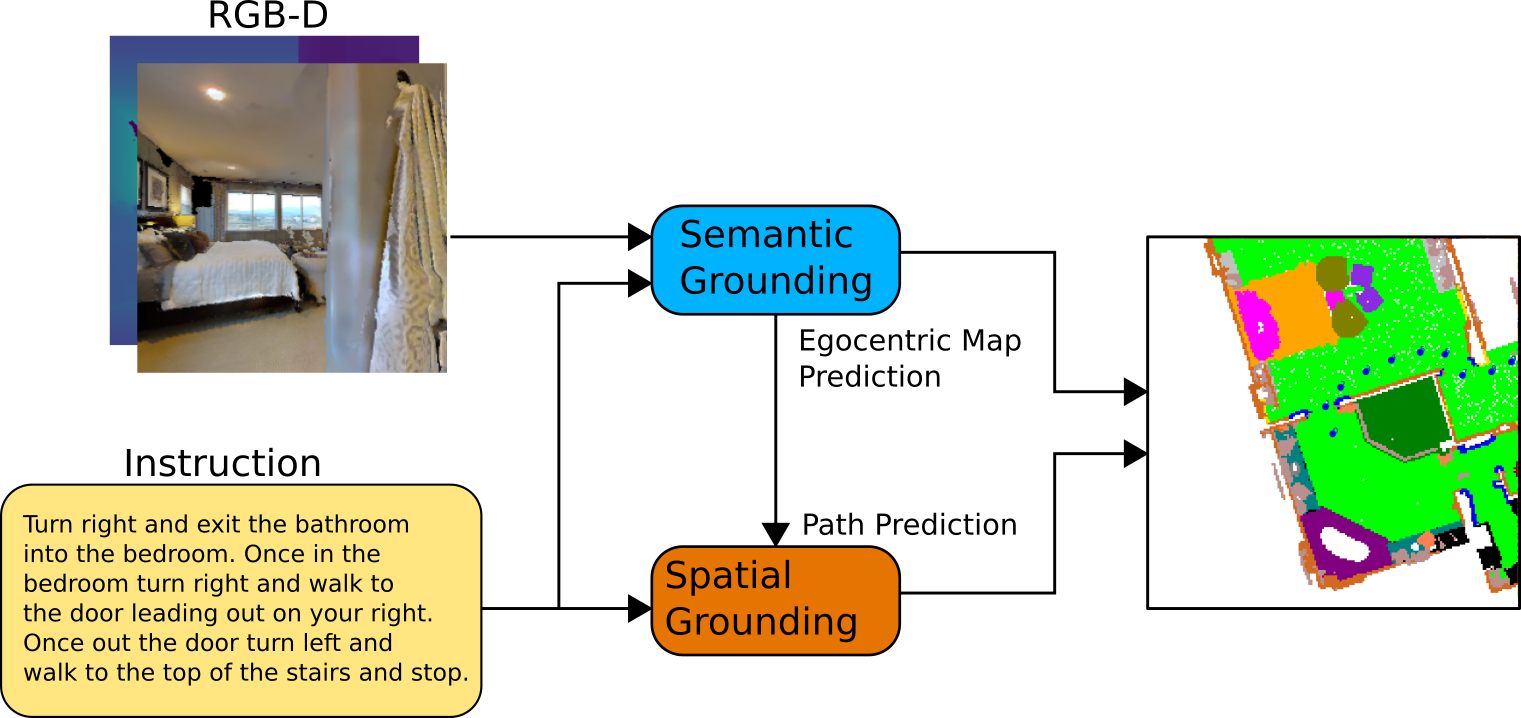}
    \caption{We approach the task of vision-and-language navigation as a two-stage procedure which learns to semantically and spatially ground the instruction on egocentric maps. }
    \label{fig:title} 
\end{figure}

We propose \textit{Cross-modal Map Learning} (CM\textsuperscript{2}), a novel navigation system for the VLN task in continuous environments, that learns a language-informed representation for both map and trajectory prediction by applying twice cross-modal attention, hence CM\textsuperscript{2}. Our method decomposes the problem in the two paths of semantic and spatial grounding as illustrated in Figure~\ref{fig:title}. First, we use a cross-modal attention network to semantically ground the instruction through an egocentric map prediction task that learns to hallucinate information outside the field-of-view of the agent. This is followed by another cross-modal attention network that is responsible for spatially grounding the instruction by learning to predict the path on the egocentric map.
Our analysis shows that through these two subtasks, the attended representations learn to focus on instruction-relevant objects and locations on the map.   

The main difference between our method and existing image-language attentional mechanisms that generate actions is that in our approach, the robot is building a cognitive map that encodes the environmental priors and follows instructions based on this map.
The motivation to use this representation is based on our finding that when the robot is given a local ground-truth (``correct'') map of the environment, then the robot outperforms all approaches on the VLN task by a large margin. This map is still local, more like a crop of the blueprint than a global map of the environment, but can still hallucinate what is behind the walls so that it can align better the map with the language instruction. 
This differentiates us from approaches such as~\cite{Chen_2021_CVPR} that first build a topological map of the environment by exploring the whole scene and then executing the task having access to a global map. 
We further argue that by learning the layout priors through cross-modal attention, we can leverage the spatial and semantic descriptions from natural language and decrease the uncertainty over the hallucinated areas.
As opposed to recent work~\cite{krantz2021waypoint} that outputs a single waypoint, we learn to predict the whole trajectory, while our waypoints are determined by the alignment between language and egocentric maps rather than the distance to the goal.





In summary, our contributions are as follows: 
\begin{itemize}[topsep=0pt,itemsep=-1ex,partopsep=1ex,parsep=1ex]
    \item A novel system for the VLN task that learns maps as an explicit intermediate representation.
    \item Semantic grounding of language to those maps by applying cross-modal attention when learning to predict semantic maps.
    \item Spatial grounding of instructions when learning to predict paths by applying cross-modal attention on semantic maps and language.
    \item An analysis over the learned representation that demonstrates the effectiveness of using egocentric maps for the VLN task.
    \item Competitive results in the VLN-CE~\cite{krantz2020beyond} dataset against current state-of-the-art methods.
\end{itemize}













\section{Related Work}

\noindent \textbf{Vision-and-Language Navigation.} The problem of instruction following for navigation has drawn significant attention in a wide range of domains. These include Google Street View Panoramas~\cite{chen2019touchdown}, simulated environments for quadcopters~\cite{blukis2018mapping}, multilingual settings~\cite{ku2020room}, interactive vision-dialogue setups~\cite{zhu2021self}, real world scenes~\cite{anderson2021sim}, and realistic simulations of indoor scenes~\cite{anderson2018vision}.
More relevant to our work is the literature on the Vision-and-Language Navigation (VLN) task initially defined in~\cite{anderson2018vision} on navigation graphs (R2R) in Matterport3D~\cite{chang2017matterport3d} dataset, and then converted for continuous environments in~\cite{krantz2020beyond} (VLN-CE). Arguably, the biggest challenges in VLN are grounding the natural language to the visual input while keeping track which part of the instruction was completed.
To address these issues, many methods rely on unstructured memory such as LSTM for visual-textual alignment~\cite{ma2019self,huang2019multi,fried2018speaker,deng2020evolving}, or have dedicated progress monitor modules~\cite{ma2019regretful,ma2019self}. Other approaches formulate instruction following as a Bayesian tracking problem~\cite{anderson2019chasing}, or learn to decompose and execute the instructions in short steps~\cite{zhu2020babywalk}. 
Another line of works~\cite{hong2021vln,qi2021road,pashevich2021episodic,hao2020towards,Chen_2021_CVPR,krantz2021waypoint,raychaudhuri2021language,guhur2021airbert,majumdar2020improving} make use of attention mechanisms and adapt powerful language models such as BERT~\cite{devlin2018bert} and transformer networks~\cite{vaswani2017attention} to the VLN task. For instance, Chen at al.~\cite{Chen_2021_CVPR} learn the association between instructions and nodes on a prebuild topological map of the environment, while Krantz et al.~\cite{krantz2021waypoint} learn to predict waypoints from panoramic images and investigate the prediction in different action spaces. In contrast to all these works, our method learns to associate the language and egocentric observations at the semantic level with 2D spatial representations followed by path prediction.

\noindent \textbf{Cross-modal attention.} The transformer architecture \cite{vaswani2017attention} has been extremely successful in language \cite{devlin2018bert}, speech \cite{dong2018speech} vision \cite{khan2021transformers} and multimodal applications \cite{hu2021transformer}. A key feature of the transformer architecture is the \textit{attention} mechanism. Cross-modal transformers have been widely used for vision-language tasks such as visual-question answering and beyond, such as joint video and language understanding \cite{sun2019videobert}. Additionally, there have been investigations into whether the multimodal transformers learn interpretable relations between the two modalities by analyzing the cross-modal attention heads, as studied in Visual BERT \cite{li2020does} and in a cross modal self attention network for referring image segmentation  \cite{ye2019cross}.
Prior works have trained cross-modal transformers in  two ways: 1) \textit{single-stream design} where the multimodal inputs (for example, word embeddings and image regions) are fed into a single transformer architecture. Examples of this are UNITER \cite{chen2020uniter}, VLBERT\cite{su2019vl}, VisualBERT \cite{li2020does}.
2) \textit{multi-stream design} where the individual modalities are encoded separately via self-attention and then a cross-modal representation is learned by the transformer. Examples of this are LXMERT \cite{tan2019lxmert}, ViLBERT \cite{lu2019vilbert}, \cite{zheng2020cross}.
In this work, we adapt the multi-stream design for vision language navigation using egocentric maps. We also investigate the cross-modal attention heads and decoder representation of the transformer for interpretable patterns.
%

\noindent \textbf{Map Prediction in Navigation.}
Modular approaches using different types of spatial representations have been successful in multiple navigation tasks, whether they focused on occupancy~\cite{gupta2017cognitive,ramakrishnan2020occupancy,chaplot2020learning,katyal2019map,georgakis2022uncertainty} or semantic map prediction~\cite{chaplot2020object,georgakis2021learning,narasimhan2020seeing,liang2020sscnav,cartillier2021semantic,georgakis2019simultaneous}. For example, Gupta et al.~\cite{gupta2017cognitive} learn a differentiable mapper for predicting top-down egocentric maps that are trained end-to-end with a differentiable planner, while Cartillier et al.~\cite{cartillier2021semantic} learn to build top-down allocentric maps from egocentric RGB-D observations. Several recent works go beyond traditional mapping and learn to predict information outside the field-of-view of the agent~\cite{ramakrishnan2020occupancy,georgakis2021learning,narasimhan2020seeing,liang2020sscnav}. The work of~\cite{ramakrishnan2020occupancy} learns to hallucinate occupancy layouts in indoor environments, while~\cite{georgakis2021learning} extends the prediction to semantic classes and uses information gain objectives to increase the performance of the predictor.
Our approach expands upon this last set of methods by presenting a language-informed model that attempts to hallucinate missing information using cues from both language and currently observed regions.

\section{Approach}

\begin{figure*}[t]
    \centering
    \includegraphics[width=0.9\linewidth]{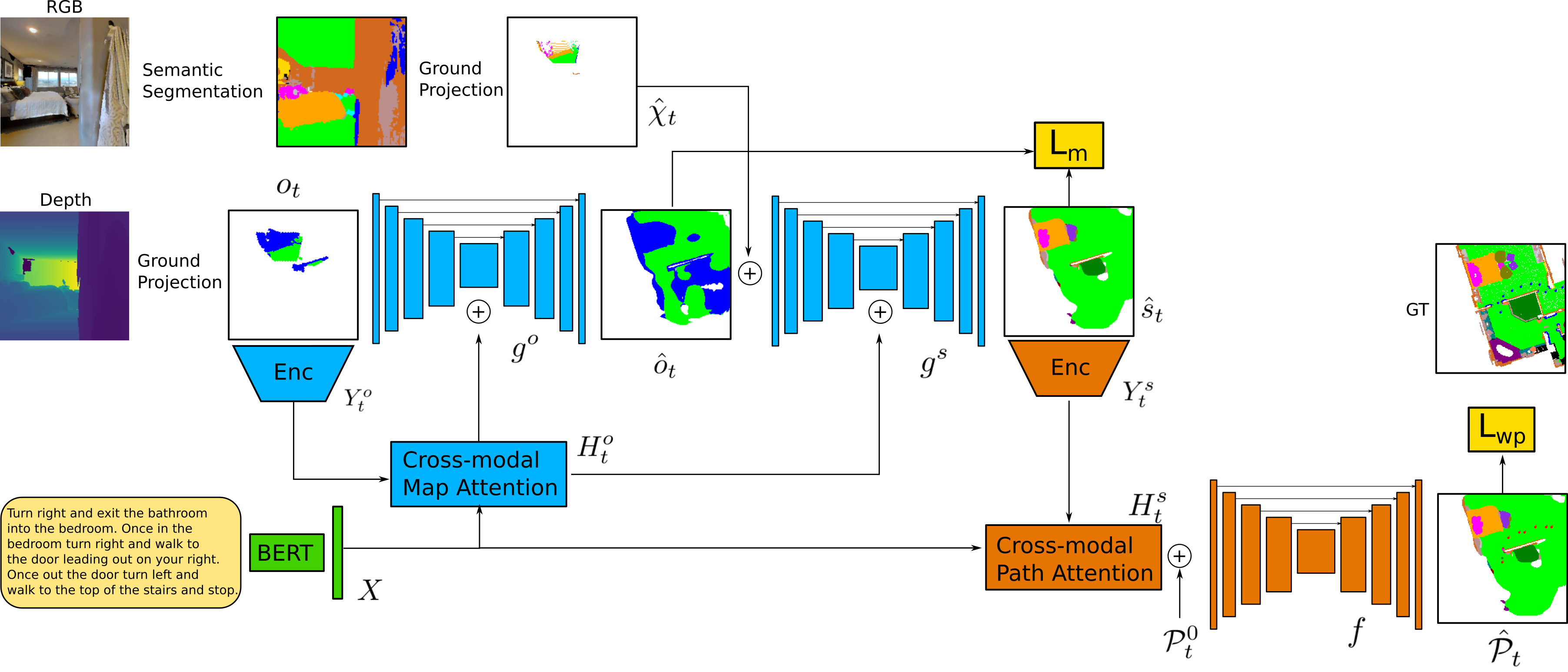}
    \caption{We propose an approach to predict egocentric semantic maps and paths described by natural language instructions. At the core of our method are two cross-modal attention modules that learn language-informed representations to facilitate both the hallucination of semantics over unobserved areas as well as the prediction of a set of waypoints that the agent needs to follow to reach the goal. The components colored in blue refer to the map prediction part (Sec.~\ref{subsec:map_attention}) of our model, the ones in orange correspond to the path prediction (Sec.~\ref{subsec:path_attention}), and the yellow boxes are the losses.}
    \label{fig:system} 
\end{figure*}

\subsection{Problem setup}
We address instruction-following navigation in indoor environments, where natural language instructions implicitly describe a specific path and goal location in the environment that an agent needs to follow. In particular, we consider the setup described in the Vision-and-Language Navigation in Continuous Environments (VLN-CE)~\cite{krantz2020beyond} that was adapted from the Room-to-Room (R2R)~\cite{anderson2018vision} dataset from pre-specified navigation graphs to continuous 3D environments.
VLN-CE uses the Habitat~\cite{savva2019habitat} simulator in the Matterport3D~\cite{chang2017matterport3d} scenes and offers more realistic settings and is much more challenging~\cite{krantz2020beyond} than the original R2R.
During a VLN-CE navigation episode, the agent has access to egocentric RGB-D observations at a resolution of $256 \times 256$ with a horizontal field-of-view of $90^\circ$. In contrast to other recent methods~\cite{krantz2021waypoint,Chen_2021_CVPR}, we assume the agent observes a frame with limited field-of-view at each time-step (not panoramas).
The action space is defined over a discrete set of actions consisting of \texttt{MOVE\_FORWARD} by $0.25m$, \texttt{TURN\_LEFT} and \texttt{TURN\_RIGHT} by $15^\circ$, and \texttt{STOP}, without actuation noise. Recently, the work of~\cite{krantz2021waypoint} demonstrated higher performance when continuous-space actions are considered, however we kept the action set discrete to remain consistent with prior work on VLN-CE.


\subsection{Overview of our approach}
We propose a method for Vision-and-Language Navigation involving path prediction over predicted semantic egocentric 2D maps.
Our argument for this approach is threefold. First, an egocentric map offers a natural representation for grounding spatial and semantic concepts from natural language instructions. Second, a VLN method should take advantage of the knowledge over semantic and spatial layouts as they offer a strong prior over possible trajectories.
Third, the language instruction provides a semantic description of a trajectory through the environment, which could be leveraged to improve map predictions.

Given the instruction, our method learns to predict the entire path defined as a set of waypoints on an egocentric local map at every step of the episode (Sec.~\ref{subsec:path_attention}). 
The agent then localizes itself on the current predicted path and chooses the following waypoint on the path as a short-term goal. This goal is then passed to an off-the-shelf local policy (DD-PPO~\cite{wijmans2019dd}) which predicts the next navigation action. We assume that we have access to ground-truth pose as provided by the simulator to facilitate DD-PPO. We note that estimating the pose from noisy sensor readings is out of the scope of this work, and point to visual odometry methods~\cite{zhao2021surprising} that can adapt DD-PPO agents to such a setting.

To obtain the egocentric map we define a language-informed two-stage semantic map predictor that learns to hallucinate the semantics in the unobserved areas (Sec.~\ref{subsec:map_attention}). An overview of our method is shown in Figure~\ref{fig:system}.
In the following two paragraphs we briefly describe the common input encoding procedures between different components of our method.





\paragraph{Instruction Encoding.} We use a pretrained Bidirectional Encoder Representations from Transformers (BERT)~\cite{devlin2018bert} model, which is a multi-layer transformer~\cite{vaswani2017attention}, to extract a feature vector for each word in the instruction. The overall feature representation for the instruction $X' \in \mathbb{R}^{M \times d'}$ is passed through a fully-connected layer to obtain the final representation $X \in \mathbb{R}^{M \times d}$, where $M$ is the number of words in the instruction, $d'=768$ is the default feature dimension of BERT, and $d=128$ is the feature dimension we use throughout our method. During training we only finetune the last layer of BERT.

\paragraph{Egocentric Map Encoding.} Our network encodes an input egocentric semantic map $s \in \mathbb{R}^{h' \times w' \times c}$ with a truncated ResNet18~\cite{he2016deep}, where $h'$, $w'$, $c$ are height, width, and the number of semantic classes, respectively as $Y=Enc\left(s\right)$. 
The ResNet18 initially produces a feature representation $Y' \in \mathbb{R}^{h \times w \times d}$ ($h=\frac{h'}{16}$, $w=\frac{w'}{16}$), which is then reshaped to $Y \in \mathbb{R}^{N \times d}$ ($N=h \times w$).
One of these modules encodes the ground projected RGB-D observations for the map predictor (Sec.~\ref{subsec:map_attention}) and a separate module is used to encode the predicted semantic map for the path predictor (Sec.~\ref{subsec:path_attention}).

\subsection{Cross-modal attention for path prediction} \label{subsec:path_attention}
The cross-modal attention for path prediction module takes as input the
instruction representation $X$ and the egocentric map encoding $Y$ and formulates the path prediction problem as a waypoint localization task.
In order to learn a grounded representation of the natural language instruction on the egocentric map, we define a cross-modal attention module following the architecture of the self-attention transformer model~\cite{vaswani2017attention}. While it is common to concatenate the representations of the two modalities and then use self-attention (such as VisualBERT~\cite{li2020does}) we follow the example of LXMERT~\cite{tan2019lxmert} and treat the egocentric map separately as the query and the instruction as the key and value. 
The idea is that during an episode the language instruction remains the same while the egocentric map changes at each time-step and is used to query the model for the path.

Specifically, given the egocentric map feature representation $Y_t^s=Enc\left(s_t\right)$ at time $t$ during a VLN episode, and the instruction features $X$, we use the scaled dot-product attention:
\begin{equation} \label{eq:matrices}
    Q=Y_t^sW_q, K=XW_k, V=XW_v
\end{equation}
\begin{equation} \label{eq:attn}
    H_t^s = Softmax \left( \frac{QK^T}{\sqrt{d}} \right) V
\end{equation}
where $W_q$, $W_k$, and $W_v \in \mathbb{R}^{d \times d}$ are learned parameter matrices, and $H_t^s \in \mathbb{R}^{N \times d}$ is the attended representation over the egocentric semantic map regions. 
In practice, this architecture~\cite{vaswani2017attention} first applies self-attention to each modality followed by the cross-modal attention. 


We define the path as a set of 2D waypoints $\{p_t^i\}_{i=1}^k$ situated on the egocentric map. The first and last waypoints always represent the starting position and the final goal position respectively. 
During training we sample the remaining waypoints from the ground-truth path with respect to the instruction.
These are used to construct ground-truth heatmaps $\mathcal{P}_t \in \mathbb{R}^{k \times u \times v}$ based on a 2D Gaussian centered at each waypoint with $\sigma=1$, where $u$, $v$ are the height and width of the heatmap. 
We predict the entire path at every time-step given the entire instruction. This can cause ambiguity with regards to the waypoint placements towards the agent's current pose, since the agent has no knowledge of the amount of the path covered at a given time-step.
In other words, if the agent is half-way through the path then the model should learn to predict both backward and forward waypoints along the path, as opposed to predicting only forward waypoints at the beginning of the episode. 
We mitigate this issue in two ways. First, the path prediction is conditioned on the starting position heatmap $\mathcal{P}_t^0$ relative to the current agent's pose. 
Second, we add an auxiliary loss that trains the model to predict a probability $\hat{\xi}_t^i$ for each waypoint whether it has already been traversed. We empirically found this auxiliary loss to help the learning process.

The waypoint predictor model is defined as an encoder-decoder UNet~\cite{ronneberger2015u} $f$, that takes as inputs the instruction attended representation of the egocentric map regions $H_t^s$ and the starting position $\mathcal{P}_t^0$:
\begin{equation}
    \hat{\mathcal{P}}_t, \hat{\xi}_t = f \left( H_t^s, \mathcal{P}_t^0 \right).
\end{equation}
We train the waypoint prediction with the following loss:
\begin{equation}
    L_{wp} = \sum_{i=1}^k b^i_t ||\hat{\mathcal{P}}_t^i - \mathcal{P}_t^i||^2_2 - \lambda_{\xi} \xi_t^i \log \hat{\xi}_t^i
\end{equation}
where $b^i_t$ is a binary indicator whether the particular waypoint $i$ is visible on the egocentric map at time $t$, and $\lambda_{\xi}$ weighs the auxiliary loss.


\subsection{Cross-modal attention for map prediction} \label{subsec:map_attention}
We design a language-informed semantic map predictor for obtaining the egocentric semantic map $s_t$ from RGB-D observations. Given the often limited field-of-view of embodied agents, we are interested in hallucinating the semantic information in regions where the agent cannot directly observe. While different versions of this procedure were attempted in the past~\cite{ramakrishnan2020occupancy,georgakis2021learning,georgakis2022uncertainty}, our key contribution is to learn the layout priors by leveraging the spatial and semantic descriptions from the instructions. 

The map prediction is defined as a semantic segmentation task over the top-down egocentric map. Our model first takes as input the depth observation which is ground-projected to an egocentric grid $o_t \in \mathbb{R}^{h' \times w' \times 3}$ containing the classes $occupied$, $free$, and $void$. For the ground-projection we first unproject the depth to a 3D point cloud using the camera intrinsic parameters and then map each 3D point to an $h' \times w'$ grid following the procedure described here~\cite{henriques2018mapnet}. Note that $o_t$ is an incomplete representation of the occupancy map around the agent, where all areas outside the field-of-view are considered unknown. 

We define a cross-modal attention module similar to the one in Sec.~\ref{subsec:path_attention}, where the feature representation $Y_t^o = Enc\left(o_t\right)$ is determined as the query, while the instruction features $X$ are used as key and value. Following Eq.~\ref{eq:matrices} and~\ref{eq:attn} (where $Y_t^s$ is replaced by $Y_t^o$) we get the attended representation $H_t^o$ over the incomplete egocentric map $o_t$.
The prediction model includes two encoder-decoder UNet~\cite{ronneberger2015u} models $g^o$, $g^s$ stacked together:
\begin{align}
    \hat{o}_t = g^o \left(o_t, H_t^o \right) && \hat{s}_t = g^s \left(\hat{o}_t, H_t^o, \hat{\chi}_t \right)
\end{align}
where $\hat{\chi}_t \in \mathbb{R}^{h' \times w' \times c}$ is a ground-projected semantic segmentation of the RGB frame.
Note that $H_t^o$ is concatenated at the bottlenecks of both $g^o$, $g^s$ models.
The model is trained with a pixel-wise cross-entropy loss on the occupancy and the semantic classes:
\begin{equation}
    L_{m} = -  \sum_{q \in (s, o)} \sum_k \sum_c q_{k,c} \log \hat{q}_{k,c}
\end{equation}
where $k$ iterates over the number of pixels in the map and $q_{k,c}$ is the ground-truth label for pixel $k$.
The ground-truth semantic maps are created from the available 3D semantic information in Matterport3D. 
The network that produces $\hat{\chi}$ is another UNet which is pre-trained separately from the rest of the model. 




\noindent \textbf{Overall learning objective.}
During training we add up all the losses from the path and map prediction modules:
\begin{equation}
    L = \lambda_{wp} L_{wp} + \lambda_m L_m
\end{equation}
where the $\lambda s$ denote the corresponding loss weights, and perform a single backward pass through the entire model.

\begin{figure*}[t]
    \centering
    \includegraphics[width=0.9\linewidth]{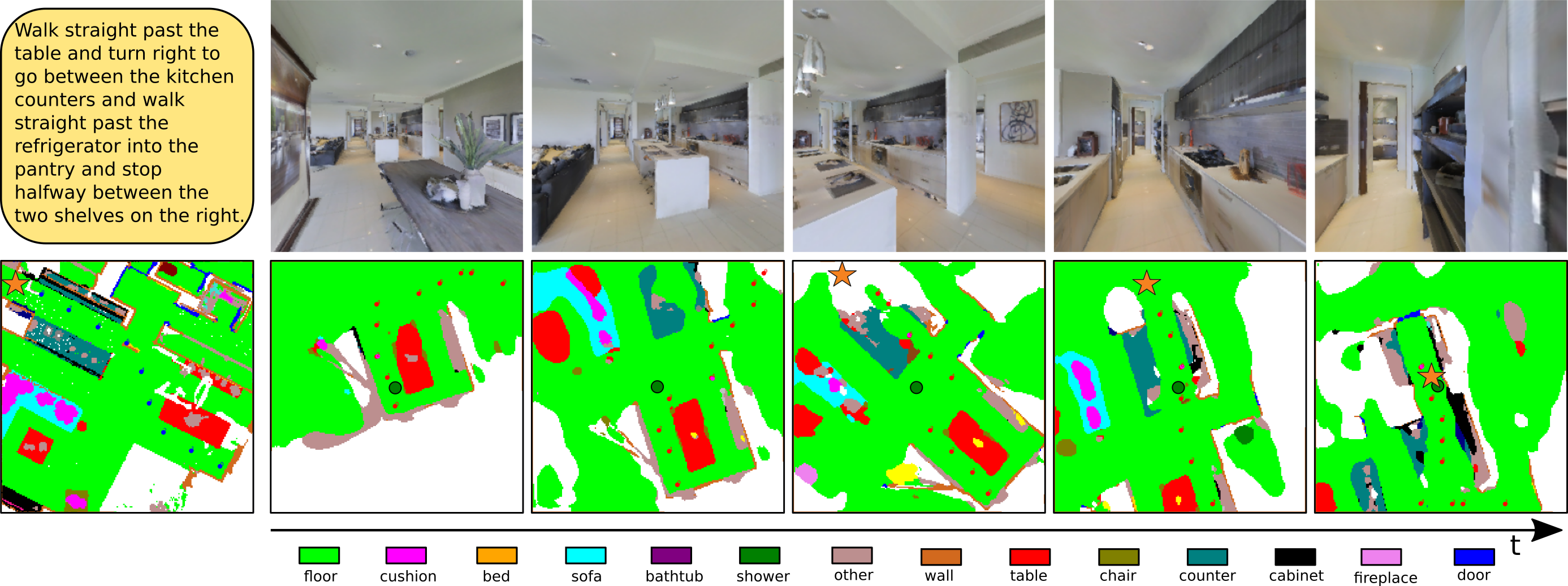}
    \caption{Navigation example using our method \textit{CM\textsuperscript{2}} on a scene from \textit{val-unseen}. The top row shows the RGB observations of the agent, while bottom shows the path prediction on the egocentric maps (the agent is in the middle looking upwards shown as the \textcolor{ao}{green circle}). The \textcolor{red}{red waypoints} represent our path prediction at the particular time-step. Observe that the goal, shown as an \textcolor{darkorange}{orange star}, is neither visible nor within the egocentric map at the beginning of the episode. The ground-truth map and path are depicted in the bottom left corner.}
    \label{fig:nav_example} 
\end{figure*}

\subsection{Controller}
The method described so far outputs the path as a set of 2D waypoints $\{p_t^i\}_{i=1}^k$ on an egocentric map from an RGB-D observation. In order to follow this path towards the goal, at each time-step we designate a waypoint as a short-term goal, following:
\begin{equation}
    \zeta = 1 + \argmin_i \Delta( \hat{p}_t^i, \varrho_t)
\end{equation}
where $\Delta$ is the euclidean distance, $\hat{p}_t^i$ corresponds to the mode of the predicted waypoint heatmap $\hat{\mathcal{P}}_t^i$, and $\varrho_t$ is the agent's pose at time $t$. 
This effectively determines the closest predicted waypoint to the agent and selects the next one in the sequence as the short-term goal $p_t^{\zeta}$. In order to reach the short-term goal, we use the off-the-shelf deep reinforcement learning model DD-PPO~\cite{wijmans2019dd} that is trained for the PointNav~\cite{anderson2018evaluation} task. 
DD-PPO receives the current depth observation and $p_t^{\zeta}$ and outputs the next navigation action for the agent. Finally, at any time during the episode the agent may decide on 
the \texttt{STOP} action when it's within a certain radius $\tau$ (m) of the final goal (last predicted waypoint) and the confidence of the goal in the predicted heatmap is above a threshold $\gamma$.


\section{Experiments}

We conduct our experiments in the VLN-CE~\cite{krantz2020beyond} dataset, which offers $16,844$ path-instruction pairs over 90 visually realistic scenes in the Matterport3D~\cite{chang2017matterport3d} dataset. We follow the typical evaluation scenario and report results in scenes which were observed (\textit{val-seen}) and not observed (\textit{val-unseen}) during training. 
An episode is considered successful if the \texttt{STOP} decision is taken within $3m$ of the goal position, and the agent has a fixed-time budget of 500 steps to complete an episode.
As mentioned before, the agent has access to egocentric RGB-D observations with a horizontal field-of-view of $90^\circ$.
We perform three sets of experiments. 
First, we compare against other methods on the VLN-CE dataset including the held-out test set of the VLN-CE challenge (Sec.~\ref{subsec:vln_eval}), followed by an ablation study (Sec.~\ref{subsec:ablation}). Finally, we provide visual examples of the learned representation (Sec.~\ref{subsec:validation}). 
We use two main variations of our method. \textit{CM\textsuperscript{2}} refers to our full pipeline that predicts both the egocentric map and path from RGB-D inputs, while \textit{CM\textsuperscript{2}-GT} refers to using the ground-truth egocentric map as input, effectively only performing path prediction.
All egocentric maps used are local $192 \times 192$ with each pixel corresponding to $5cm \times 5cm$. The map covers a square 9.6 meters on a side, leaving most of the scene unobserved.
We provide code, trained models and instructions to reproduce our results: \url{https://github.com/ggeorgak11/CM2}. Implementation details along with additional experimental results are included in the appendix.

\begin{table*}[t]
    \centering
    \scalebox{0.90}{
    \begin{tabular}{lcccccccccc}
    \hline
    & \multicolumn{5}{c}{Val-Seen} & \multicolumn{5}{c}{Val-Unseen} \\
    \cline{2-6} \cline{7-11}
     & TL $\downarrow$ & NE $\downarrow$ & OS $\uparrow$ & SR $\uparrow$ & SPL $\uparrow$ & TL $\downarrow$ & NE $\downarrow$ & OS $\uparrow$ & SR $\uparrow$ & SPL $\uparrow$ \\
    \hline
    Seq2Seq+PM+DA+Aug~\cite{krantz2020beyond} & 9.37 & 7.02 & 46.0 & 33.0 & 31.0 & 9.32 & 7.77 & 37.0 & 25.0 & 22.0 \\
    AG-CMTP*~\cite{Chen_2021_CVPR} & - & 6.60 & \textbf{56.2} & 35.9 & 30.5 & - & 7.9 & 39.2 & 23.1 & 19.1\\
    R2R-CMTP*~\cite{Chen_2021_CVPR} & - & 7.10 & 45.4 & 36.1 & 31.2 & - & 7.9 & 38.0 & 26.4 & 22.7 \\
    CMA+PM+DA+Aug~\cite{krantz2020beyond} & 9.26 & 7.12 & 46.0 & 37.0 & 35.0 & 8.64 & 7.37 & 40.0 & 32.0 & 30.0 \\
    WPN-DD*~\cite{krantz2021waypoint} & \textbf{9.11} & 6.57 & 44.0 & 35.0 & 32.0 & \textbf{8.23} & 7.48 & 35.0 & 28.0 & 26.0 \\
    LAW~\cite{raychaudhuri2021language} & 9.34 & 6.35 & 49.0 & 40.0 & \textbf{37.0} & 8.89 & \textbf{6.83} & \textbf{44.0} & \textbf{35.0} & \textbf{31.0} \\    
    CM\textsuperscript{2} (Ours) & 12.05 & \textbf{6.10} & 50.7 & \textbf{42.9} & 34.8 & 11.54 & 7.02 & 41.5 & 34.3 & 27.6\\ 
    \hline
    WPN-CC*~\cite{krantz2021waypoint} & 10.29  & 6.05  & 51.0 & 40.0 & 35.0 & 10.62 & 6.62 & 43.0 & 36.0 & 30.0 \\
    HPN-C*~\cite{krantz2021waypoint} & 8.71 & 5.17 & 53.0 & 47.0 & 45.0 & 7.71 & 6.02 & 42.0 & 38.0 & 36.0 \\
    \hline
    CM\textsuperscript{2}-GT (Ours) & 12.60 & 4.81 & 58.3 & 52.8 & 41.8 & 10.68 & 6.23 & 41.3 & 37.0 & 30.6 \\ 
    \hline 
    \end{tabular}
    }
    \caption{Evaluation on VLN-CE dataset. All methods marked with * use panoramic images.
    CM\textsuperscript{2}-GT is the same as CM\textsuperscript{2}, but uses ground truth local maps, rather than predicting them.
    HPN-C and WPN-CC use a more expressive action space than the rest of the methods.  
    AG-CMTP and R2R-CMPT allow the agent to explore each scene before the experiment begins.
    Our method is the most successful on \textit{val-seen} while it is competitive on \textit{val-unseen}.
    }
    \label{tab:vln_res}
\end{table*}

\subsection{VLN-CE Evaluation} \label{subsec:vln_eval}
Here we evaluate the performance of our method on the continuous vision-and-language navigation task against current state-of-the-art methods. The metrics reported are the following: Trajectory length \textbf{TL}~(m), navigation error from goal \textbf{NE}~(m), oracle success rate \textbf{OS}~(\%), success rate \textbf{SR}~(\%), and success weighted by path length \textbf{SPL}~(\%). More details on these metrics can be found in~\cite{anderson2018evaluation,anderson2018vision}.

We compare our method against the following works:

\noindent \textbf{Krantz et al.}~\cite{krantz2020beyond}: Two baselines are used from here. First, \textit{Seq2Seq+PM+DA+Aug} is a simple sequence-to-sequence baseline that uses a recurrent policy to predict the action directly from the visual observations. Second, \textit{CMA+PM+DA+Aug} utilizes cross-modal attention between instruction and RGB-D observations. Both methods use off-the-shelf techniques for Progress Monitor (PM), DAgger (DA), and synthetic data augmentation.

\noindent \textbf{Chen et al.}~\cite{Chen_2021_CVPR}: This work uses cross-modal attention between the instruction and a topological map to compute a global navigation plan. To construct the topological map, the authors assume the agent can explore the environment before the execution of the navigation episode. Each node in the topological map corresponds to a panoramic image. We compare against \textit{AG-CMTP} and \textit{R2R-CMTP} which use the method's generated map and the maps from the Room2Room~\cite{anderson2018evaluation} dataset respectively. 

\noindent \textbf{Raychaudhuri et al.}~\cite{raychaudhuri2021language}: This method (\textit{LAW}) updates the training setup of \textit{CMA+PM+DA+Aug}~\cite{krantz2020beyond} by adjusting the supervision to use the nearest waypoint on the path rather than the goal location. 

\noindent \textbf{Krantz et al.}~\cite{krantz2021waypoint}: We compare against the Waypoint Prediction Network (\textit{WPN}) and the Heading Prediction Network (\textit{HPN}) which are end-to-end models that predict relative waypoints directly from natural language instructions and panoramic RGB-D inputs. The models differ with respect to the waypoint prediction space. \textit{WPN-CC} considers continuous values for distance and direction, \textit{WPN-DD} considers discrete values, and \textit{HPN-C} uses a constant value for distance and continuous for direction. Our method is analogous to \textit{WPN-DD}, since our waypoint prediction is on the discrete 2D space of maps. Investigating more expressive waypoint prediction spaces is out of the scope of our work.

Quantitative results are shown in Table~\ref{tab:vln_res} and a navigation example can be seen in Figure~\ref{fig:nav_example}. On \textit{val-seen} our method \textit{CM\textsuperscript{2}} outperforms all other baselines except \textit{WPN-CC} and \textit{HPN-C} (which use more expressive waypoint prediction spaces) on navigation error and success rate, while it is competitive on SPL. In particular, we show better results than \textit{WPN-DD} which uses panoramic images (4$\times$ larger field-of-view), and was trained on 200M steps of experience (285$\times$ more data)~\cite{krantz2021waypoint}.
This is a characteristic of end-to-end methods that need to learn all navigation components such as mapping, planning, and control in a single network and thus require large amounts of data. In contrast, aligning language to egocentric maps proves to be much more sample efficient, as our model was trained with only 0.7M training samples.
Regarding our comparison to~\cite{Chen_2021_CVPR}, \textit{AG-CMPT} performs better only on oracle success rate, while our \textit{CM\textsuperscript{2}} method has a noticeably higher success rate. 
However, this baseline has a prior scene exploration phase, which is not counted in the task step limit, that acquires knowledge of scene topology to use during the navigation episode.
In comparison, our \textit{CM\textsuperscript{2}-GT}, which also has knowledge over the map, performs better on all metrics.
We are also competitive against \textit{CMA+PM+DA+Aug} and \textit{LAW} that use cross-modal attention mechanisms between the instruction and the RGB-D frames.
The latter also employs a more sophisticated reward function that forces the agent to stay on the path and trains on an augmented dataset with over ten times as many trajectories.
We outperform both in success rate on \textit{val-seen} and we have almost the same performance with \textit{LAW} on \textit{val-unseen}.
Finally, when the input to our method is the ground-truth egocentric semantic map (\textit{CM\textsuperscript{2}-GT}) we observe a significant increase in success rate in \textit{val-seen}. Although the map is local and the goal location is usually not visible, this performance gain further justifies our choice of using cross-modal attention on egocentric maps.  



\begin{table}[t]
    \centering
    \scalebox{0.9}{
    \begin{tabular}{lccccc}
        \hline
         Team Name & TL & NE & OS & SR & SPL \\
        \hline
        CWP-VLNBERT* & 13.3 & 5.9 & 51 & 42 & 36 \\
        CWP-CMA* & 11.9 & 6.3 & 49 & 38 & 33 \\
        WaypointTeam* & 8.0 & 6.6 & 37 & 32 & 30 \\
        CM\textsuperscript{2} & 13.9 & 7.7 & 39 & 31 & 24 \\
        TJA* & 10.4 & 8.1 & 42 & 29 & 27 \\
        VIRL\_Team & 8.9 & 7.9 & 36 & 28 & 25 \\
        \hline
    \end{tabular}
    }
    \caption{Results on the VLN-CE challenge leaderboard. Methods marked with * use either panoramic images and/or a non-standard action space.}
    \label{tab:challenge}
\vspace{-2mm}
\end{table}

\paragraph{VLN-CE Leaderboard} We submitted our CM\textsuperscript{2} on the held-out test-unseen set containing 3.4K episodes in unseen environments used for the VLN-CE challenge. Table~\ref{tab:challenge} shows the leaderboard as accessed on Mar 8th 2022. 
Our method is leading in terms of OS, SR, and NE among those that use standard observation (no panoramas) and action spaces (discrete), and is 4th overall on OS SR, and NE.



\begin{table}[t]
    \centering
    \scalebox{0.9}{
    \begin{tabular}{lccc}
        \hline
         & IoU (\%) & F1 (\%) & PCW (\%) \\
        \hline
        CM\textsuperscript{2}-w/o-MapAttn & 21.2 & 33.2 & 71.1 \\
        CM\textsuperscript{2} & \textbf{28.3} & \textbf{42.2} & \textbf{76.5} \\
        \hline
    \end{tabular}
    }
    \caption{Effect of map attention on map and waypoint prediction.}
    \label{tab:map_attention}
\vspace{-2mm}
\end{table}

\begin{table}[t]
    \centering
    \scalebox{0.9}{
    \begin{tabular}{lccccc}
        \hline
        Val-Seen & TL & NE & OS & SR  & SPL \\
        \hline
        CM\textsuperscript{2}-GT, $\tau=1.5$ & \textbf{10.18} & 5.01 & 53.6 & 49.5 & 45.1 \\
        CM\textsuperscript{2}-GT, $\tau=1.0$ & 11.48 & 4.94 & 56.4 & 51.9 & 43.8 \\
        CM\textsuperscript{2}-GT, $\tau=0.5$ & 12.60 & 4.81 & 58.3 & 52.8 & 41.8 \\
        CM\textsuperscript{2}-GT-384, $\tau=0.5$ & 12.89 & \textbf{4.52} & \textbf{66.4} & \textbf{58.4} & \textbf{46.7} \\
        \hline
    \end{tabular}
    }
    \caption{Effect of map size and stop distance threshold on VLN.}
    \label{tab:map_size_stop_ablation}
\vspace{-2mm}
\end{table}

\subsection{Ablation Study} \label{subsec:ablation}

\begin{figure}[t]
    \centering
    \includegraphics[width=1\linewidth]{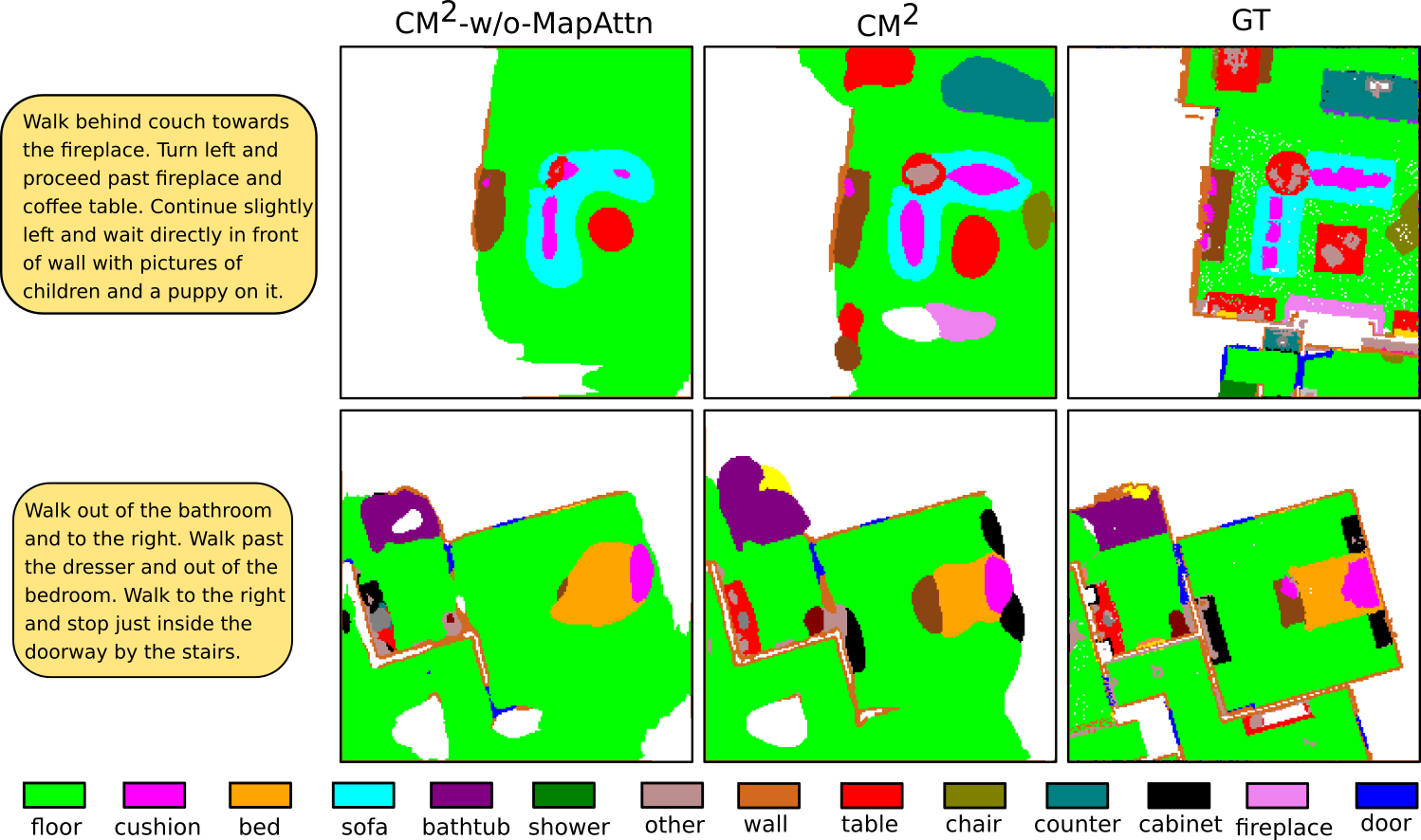}
    \caption{Semantic map predictions with and without cross-modal map attention.}
    \label{fig:map_attention} 
\vspace{-2mm}
\end{figure}

In this experiment we provide an analysis over our model and aim to answer the following questions:

\noindent \textbf{How important is the cross-modal map attention?} 
The cross-modal map attention, shown in Figure~\ref{fig:system}, is the attention module that learns the semantic grounding and influences the semantic map prediction.
We are interested in quantifying its contribution towards the map and path prediction and define the baseline \textit{CM\textsuperscript{2}-w/o-MapAttn} that does not include the cross-modal map attention module and therefore is not aware of the language instruction. We compare against our method \textit{CM\textsuperscript{2}} on the popular semantic segmentation metrics of Intersection over Union (IoU) and F1 score, and on the Percentage of Correct Waypoints (PCW) that evaluates the quality of the path prediction. PCW counts a predicted waypoint as correct if it is within $1.92m$ (on the $192 \times 192$ maps) of the ground-truth waypoint. Results are reported in Table~\ref{tab:map_attention}. \textit{CM\textsuperscript{2}} has higher performance on IoU, F1, and PCW by $7.1\%$, $9.0\%$, and $5.4\%$ respectively. 
These results show that the cross-modal map attention extracts useful information from language that improves the prediction of the semantic map and the path.
Examples over map predictions are shown in Figure~\ref{fig:map_attention}.

\noindent \textbf{What is the effect of the stop decision threshold?} We vary the stop decision distance threshold $\tau$ (m) used by the controller and observe the performance on the VLN-CE metrics in Table~\ref{tab:map_size_stop_ablation}. This experiment is carried out on \textit{val-seen} using \textit{CM\textsuperscript{2}-GT}. When $\tau=1.5$, success rate drops by $3.3\%$ because the agent chooses to stop more aggressively thus it is more likely to choose \texttt{STOP} outside the goal radius. 
On the other hand, SPL gained $3.3\%$ since stopping earlier reduces the path length. This result signifies a trade-off between success rate (SR) and SPL based on the value of $\tau$ that can adjust the agent's behavior.

\noindent \textbf{What is the effect of egocentric map size?} All experimental evaluation of our work (\textit{CM\textsuperscript{2}}, \textit{CM\textsuperscript{2}-GT}) uses $192 \times 192$ egocentric maps. Given that each cell in the map corresponds to $5cm \times 5cm$, this translates to a distance from the center of the map (where the agent is situated) to each side of 4.8m. With the mean euclidean distance between the start position and goal being around 8m across \textit{val-seen} and \textit{val-unseen} episodes, this means that for the majority of the episodes the goal is not located within the egocentric map at the beginning. In order to see how much this affects performance, we train our path predictor again \textit{CM\textsuperscript{2}-GT-384} with maps of size $384 \times 384$ (9.6m between the agent and the sides of the map) and compare to our original method in Table~\ref{tab:map_size_stop_ablation}. Doubling the map size increases SR by $5.6\%$, OS by $8.1\%$, and SPL by $4.9\%$ demonstrating that the larger maps have significant impact on the navigation performance.



\begin{figure}[t]
    \centering
    \includegraphics[width=0.85\linewidth]{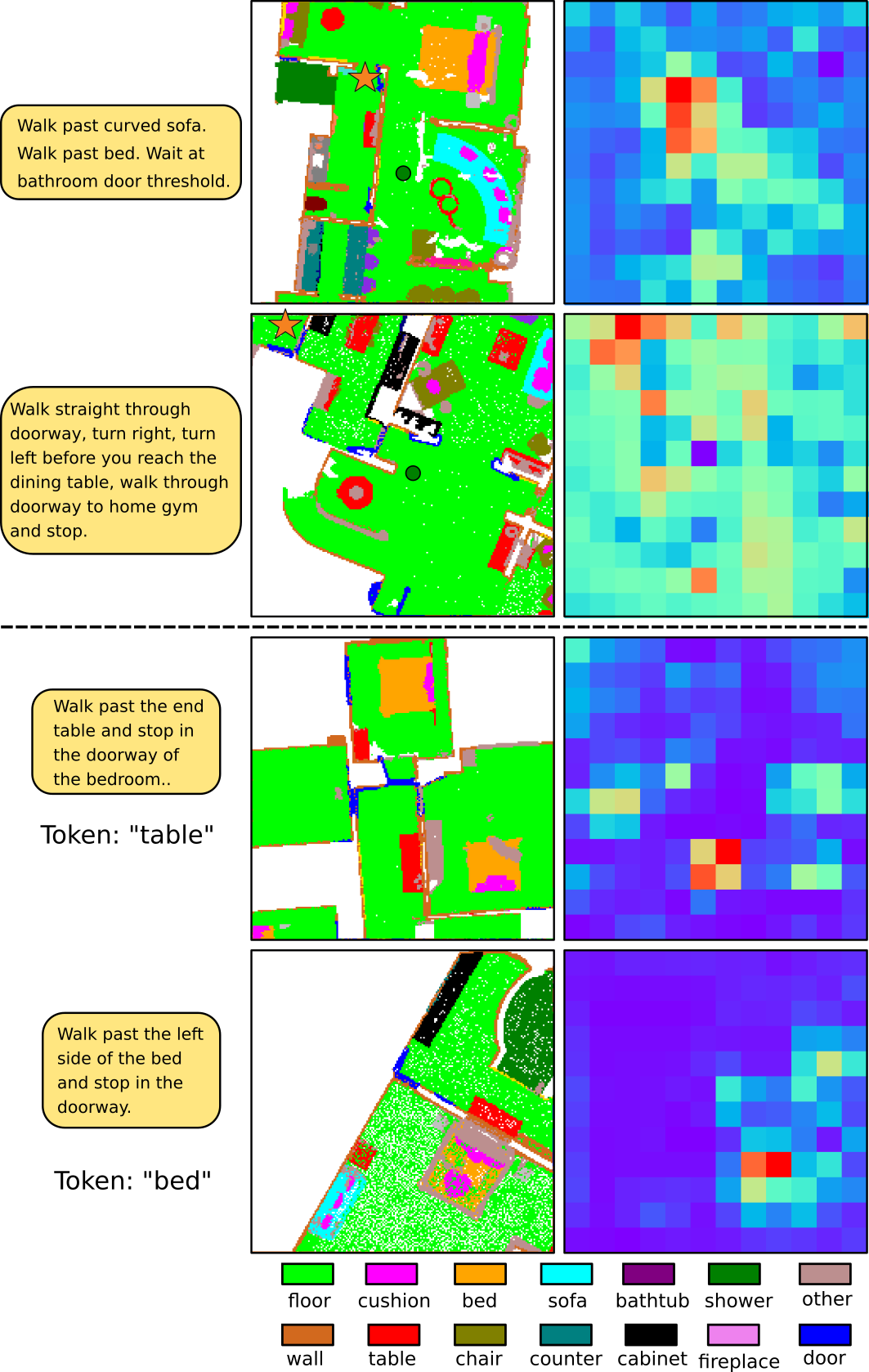}
    \caption{ Top: Visualization of attention decoder output $H_t^s$ which tends to focus on areas corresponding to goal locations. The agent's location is denoted with a \textcolor{ao}{green circle} and the goal with an \textcolor{darkorange}{orange star}. Bottom: Cross-modal attention between map and specific word tokens. }
    \label{fig:spatial_grounding}
\vspace{-2mm}
\end{figure}

\subsection{Validation of Semantic and Spatial Grounding} \label{subsec:validation}

Finally, we provide evidence that the learned representations can be semantically and spatially grounded on the egocentric maps. 
Specifically, we visualize (Figure~\ref{fig:spatial_grounding}) two feature representations from the cross-modal path attention module: 1) The attention decoder output $H_t^s \in \mathbb{R}^{N \times d}$ which we max-pool over the feature dimension $d$ and reshape $N$ back to its encoded map dimensions of $h \times w$ to get a spatial heatmap. This representation is shown to focus around goal locations and along paths.
2) The cross-modal attention ($Softmax \left( \frac{QK^T}{\sqrt{d}} \right)$) between the map regions and the words in the instruction with dimensionality $N \times M$ from which we can visualize the attention heatmap for a specific word token over the map. This demonstrates that the cross-modal attention learns to associate instruction tokens to semantic objects on the map.



\section{Conclusion}

We presented a new method for the Language-and-Navigation task that solves the problem by first predicting the egocentric semantic map and then estimating the trajectory, defined by the instruction, on the 2D map.
This is facilitated by two cross-modal attention modules that learn to semantically and spatially ground the natural language on the egocentric map.
We showcased the effectiveness of our method with competitive results on the VLN-CE dataset and demonstrated that grounding the language on the maps allows for good VLN performance with a fraction of the data that the end-to-end methods require.
Furthermore, we qualitatively show that our method learns meaningful intermediate representations.

\paragraph{Acknowledgements.} Research was sponsored by the Army Research Office and was accomplished under Grant Number W911NF-20-1-0080, as well as by the ARL DCIST CRA W911NF-17-2-0181, NSF TRIPODS 1934960, and NSF CPS 2038873 grants.

{\small
\bibliographystyle{ieee_fullname}
\bibliography{egbib}
}

\appendix
\section{Appendix}

Here we provide the following additional material:
\begin{enumerate}
    \item Discussion on societal impact and limitations.
    \item Implementation details.
    \item Analysis of path prediction learning with regards to auxiliary loss and start position input.
    \item Analytical results over semantic map prediction to assess the contribution of cross-modal map prediction.
    \item Additional results on the effect of stop decision threshold.
    \item Additional qualitative navigation results and visualizations of the learned attention representations.
\end{enumerate}

\subsection{Societal Impact and Limitations}

\paragraph{Potential negative societal impact.}
Our current method is trained on scenes from Matterport3D which contains scans of homes from North America and Europe. Since we do not model out-of-distribution scenarios, deploying our method in safety critical situations such as rescue operations or hospitals could have negative outcomes. Furthermore, house layouts strongly correlate with regions of the world and with socio-economic factors, making it likely that agents using our algorithm will underperform when deployed in other parts of the world or in poor or minority houses which are frequently underrepresented in datasets.

\paragraph{Limitations.}
While our approach achieves results comparable with the state of the art, we acknowledge that there is much room for improvement.We would like to point out three limitations of our method. First, since we predict the path from the semantic map, we are not utilizing information from the instructions that describe object attributes such as color, (i.e.,  ``brown table", ``red table"). This can be important in situations where we need to distinguish between two instances of the same category.
Second, we depend on the pretrained BERT representation, after fine-tuning its final layer, to provide all relevant information about the instruction.  We do not use any explicit language representation, which could allow for better decomposition of instructions.
Third, our method is limited by size of the local egocentric map. We cannot spatially ground information to locations outside of the local map, and while increasing the size of the local map
can significantly improve performance, it is also computationally expensive.

\begin{table}[t]
    \centering
    \scalebox{0.87}{
    \begin{tabular}{lccccc}
        \hline
         & TL & NE & OS & SR & SPL \\
        \hline
        CM\textsuperscript{2}-GT, w/o $\mathcal{P}_t^0$, $\lambda_{\xi}=0$ & \textbf{9.37} & 6.80 & 32.9 & 29.3 & 22.2 \\
        CM\textsuperscript{2}-GT, w/o $\mathcal{P}_t^0$ & 10.62 & 6.18 & 38.4 & 34.3 & 26.5 \\
        CM\textsuperscript{2}-GT, $\lambda_{\xi}=0$ & 12.61 & 5.04 & 54.3 & 49.1 & 39.0 \\
        CM\textsuperscript{2}-GT & 12.60 & \textbf{4.81} & \textbf{58.3} & \textbf{52.8} & \textbf{41.8} \\
        \hline
    \end{tabular}
    }
    \caption{Analysis of our path prediction strategy demonstrating the contributions of $\mathcal{P}_t^0$ and the auxiliary loss using navigation metrics on \textit{val-seen} set.}
    \label{tab:auxiliary_loss_ablation}
\end{table}

\begin{figure*}[t]
    \centering
    \includegraphics[width=0.8\linewidth]{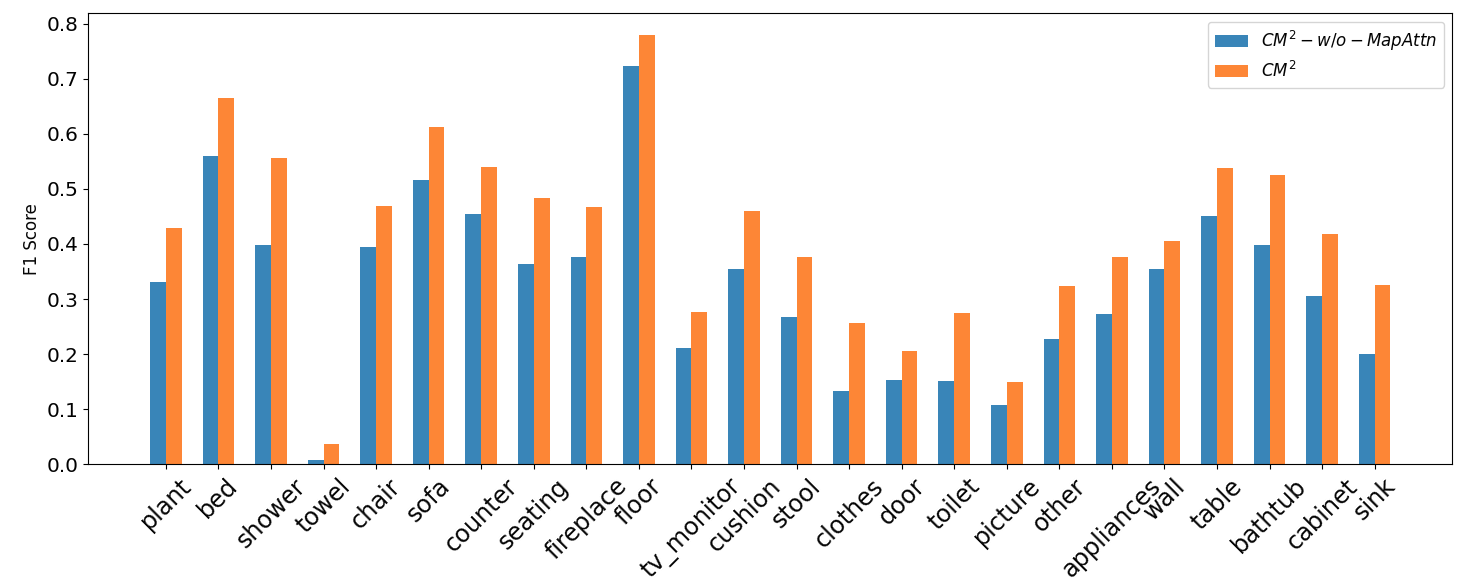}
    \caption{Per-class semantic map predictions with and without cross-modal map attention. Performance gains are more noticeable for object categories over floor and wall.}
    \label{fig:map_attention_per_object}
\end{figure*}

\begin{figure}[t]
    \centering
    \includegraphics[width=0.8\linewidth]{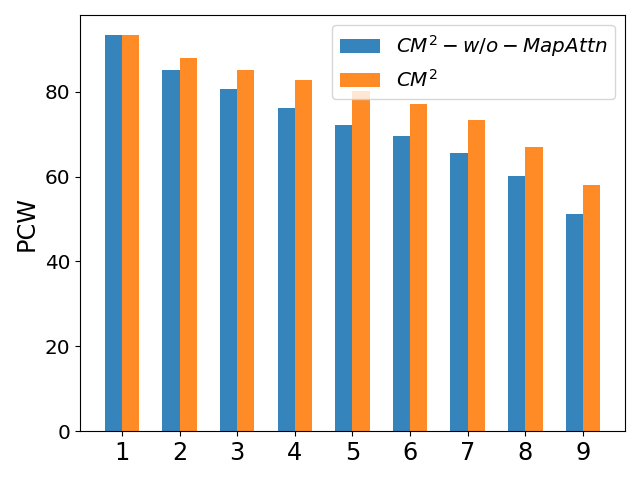}
    \caption{Per-waypoint path prediction results with and without cross-modal map attention. Waypoint 9 corresponds to the goal, while waypoint 0 is used as input to our method.}
    \label{fig:pck_per_waypoint}
\end{figure}

\subsection{Implementation details}
Our method is implemented in PyTorch~\cite{paszke2017automatic}. The UNet~\cite{ronneberger2015u} models used in our method have four encoder and four decoder convolutional blocks with skip connections. The entire model is trained with the Adam optimizer and a learning rate of 0.0002. During training all $\lambda$s are equal to 1. The training data for both the map and waypoint prediction were sampled from the ground-truth paths provided in VLN-CE train split. We used around 700K examples to train \textit{CM\textsuperscript{2}} and around 500K to train \textit{CM\textsuperscript{2}-GT}. The semantic segmentation that produces $\hat{\chi}$ is another UNet which we pre-trained separately from the rest of the model on RGB observations from the Matterport3D scenes. The egocentric map and waypoint heatmap dimensions are $h'=w'=192$ and $u=v=24$ respectively. Each pixel in the egocentric map corresponds to physical dimensions of $5cm \times 5cm$. We use $k=10$ waypoints and $c=27$ semantic classes from the original 40 categories of Matterport3D. For the controller we define stop distance threshold $\tau=0.5$ and goal confidence threshold $\gamma=0.6$. 
Our method does not use any recurrence or an implicit state representation so the map and path predictions are temporally independent. However, during a navigation episode we maintain a global occupancy map using the ground-projected depth $o_t$ that is registered using Bayesian updates. The input to the model is an egocentric crop from this global map, so the agent is aware of previously observed occupancy. 

\subsection{Analysis of path prediction learning}
We investigate the contribution of certain choices we made to mitigate the ambiguity over waypoint placements during path prediction learning as discussed in section 3.3 of the main paper. In particular, we train the following variants of our CM\textsuperscript{2}-GT model: 1) without using the starting position heatmap $\mathcal{P}_t^0$ as input, 2) without the auxiliary loss for predicting whether a waypoint has been traversed ($\lambda_{\xi}=0$), and 3) without $\mathcal{P}_t^0$ and $\lambda_{\xi}=0$. The variants are evaluated against our proposed approach on \textit{val-seen} using the navigation metrics from section 4.1 of the main paper (Table~\ref{tab:auxiliary_loss_ablation}). We observe that without the auxiliary loss success rate drops by $3.7\%$, while not using the starting position further decreases success rate by $18.5\%$. The worst performance by far is recorded when both are not utilized. The results justify our choices and suggest the importance of anchoring the prediction of the entire path to a starting location in the egocentric map, complemented by an auxiliary objective that forces the model to predict its current position on the path.

\begin{table*}[t]
    \centering
    \begin{tabular}{lcccccccccc}
        \hline
        & \multicolumn{5}{c}{Val-Seen} & \multicolumn{5}{c}{Val-Unseen} \\
        \cline{2-6} \cline{7-11}
         & TL & NE & OS & SR & SPL & TL & NE & OS & SR  & SPL \\
        \hline
        CM\textsuperscript{2}, $\tau=1.5$ & \textbf{9.54} & 6.06 & 42.4 & 38.8 & 34.6 & \textbf{9.07} & \textbf{7.01} & 35.2 & 31.3 & 27.7 \\
        CM\textsuperscript{2}, $\tau=1.0$ & 10.72 & \textbf{5.88} & 49.2 & 42.6 & \textbf{35.9} & 10.04 & 7.09 & 39.0 & 33.3 & \textbf{27.9} \\
        CM\textsuperscript{2}, $\tau=0.5$ & 12.05 & 6.10 & \textbf{50.7} & \textbf{42.9} & 34.8 & 11.53 & 7.02 & \textbf{41.5} & \textbf{34.3} & 27.6 \\
        \hline
    \end{tabular}
    \caption{Additional results on the effect of stop distance threshold on VLN.}
    \label{tab:stop_ablation_supp}
\end{table*}

\subsection{Analytical results for cross-modal map attention}
In section 4.2 of the main paper we investigated the importance of the cross-modal map attention component by comparing our approach to the baseline CM\textsuperscript{2}-w/o-MapAttn that is unaware of the language instruction during map prediction. Here, we show additional per-class and per-waypoint results over F1 score (Figure~\ref{fig:map_attention_per_object}) and PCW (Figure~\ref{fig:pck_per_waypoint}) respectively. First, in Figure~\ref{fig:map_attention_per_object} we observe that the model trained with the cross-modal map attention (CM\textsuperscript{2}) performs better on all semantic categories against the baseline. Furthermore, the performance gain is more pronounced over object categories (e.g., toilet $12.4\%$, sink $12.6\%$) as opposed to semantic classes referring to the structure of the scene (e.g., floor $5.6\%$, wall $5.1\%$).
This reinforces our initial hypothesis that the attention component is able to pick semantic cues from the instruction and improve the map prediction.
Additionally, in Figure~\ref{fig:pck_per_waypoint} we demonstrate path prediction results over individual waypoints (1-9). Waypoint 0 is omitted since it is used as input to our method, while waypoint 9 corresponds to the goal location. As expected, waypoints earlier in the path have larger PCW. However, an interesting observation is that the gain in performance increases for waypoints closer to the goal rather than in the beginning of the path, thus demonstrating that improved map prediction is crucial for predicting waypoints far from the starting position.

\begin{figure*}[t]
    \centering
    \includegraphics[width=0.8\linewidth]{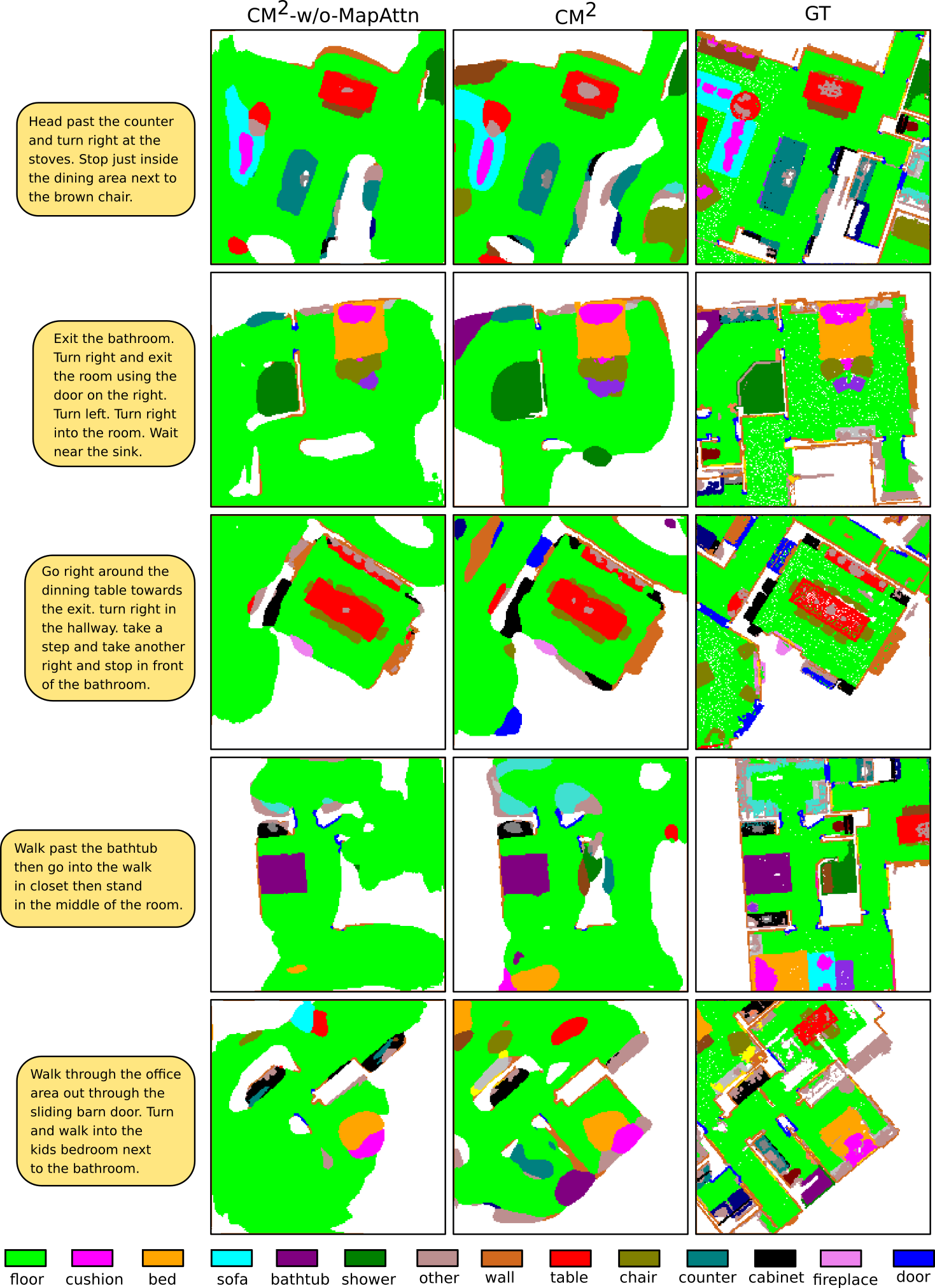}
    \caption{Semantic map predictions with and without cross-modal map attention.}
    \label{fig:map_attention_supp}
\end{figure*}

For additional qualitative comparisons of semantic map predictions between the baseline and our approach see Figure~\ref{fig:map_attention_supp}.

\subsection{Additional results on effect of stop distance threshold.}
We repeat the experiment presented in section 4.2 of the main paper regarding the effect of the stop distance threshold on the VLN task using our CM\textsuperscript{2} (no GT map) agent on both val-seen and val-unseen splits. In Table~\ref{tab:stop_ablation_supp} we observe a similar trend as that shown in Table 4 of the main paper. Success rate is higher when $\tau$ is low, because the agent takes the stop action more cautiously, while trajectory length is best when $\tau$ is high.

\subsection{Additional visualizations}
\begin{figure}[t]
    \centering
    \includegraphics[width=0.85\linewidth]{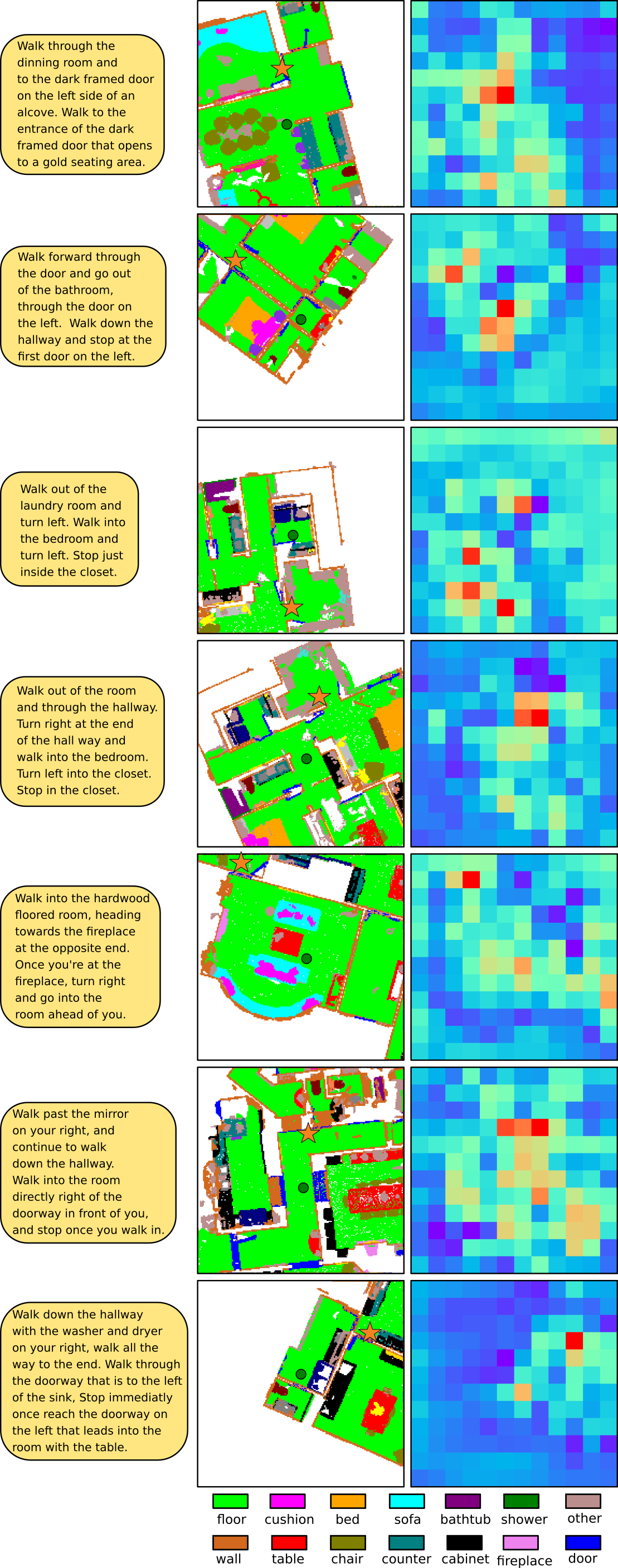}
    \caption{Visualization of attention decoder output $H_t^s$ that focuses on areas around goal locations and along paths. The agent's location is denoted with a \textcolor{ao}{green circle} and the goal with an \textcolor{darkorange}{orange star}.}
    \label{fig:spatial_grounding_supp}
\end{figure}

\begin{figure}[t]
    \centering
    \includegraphics[width=0.85\linewidth]{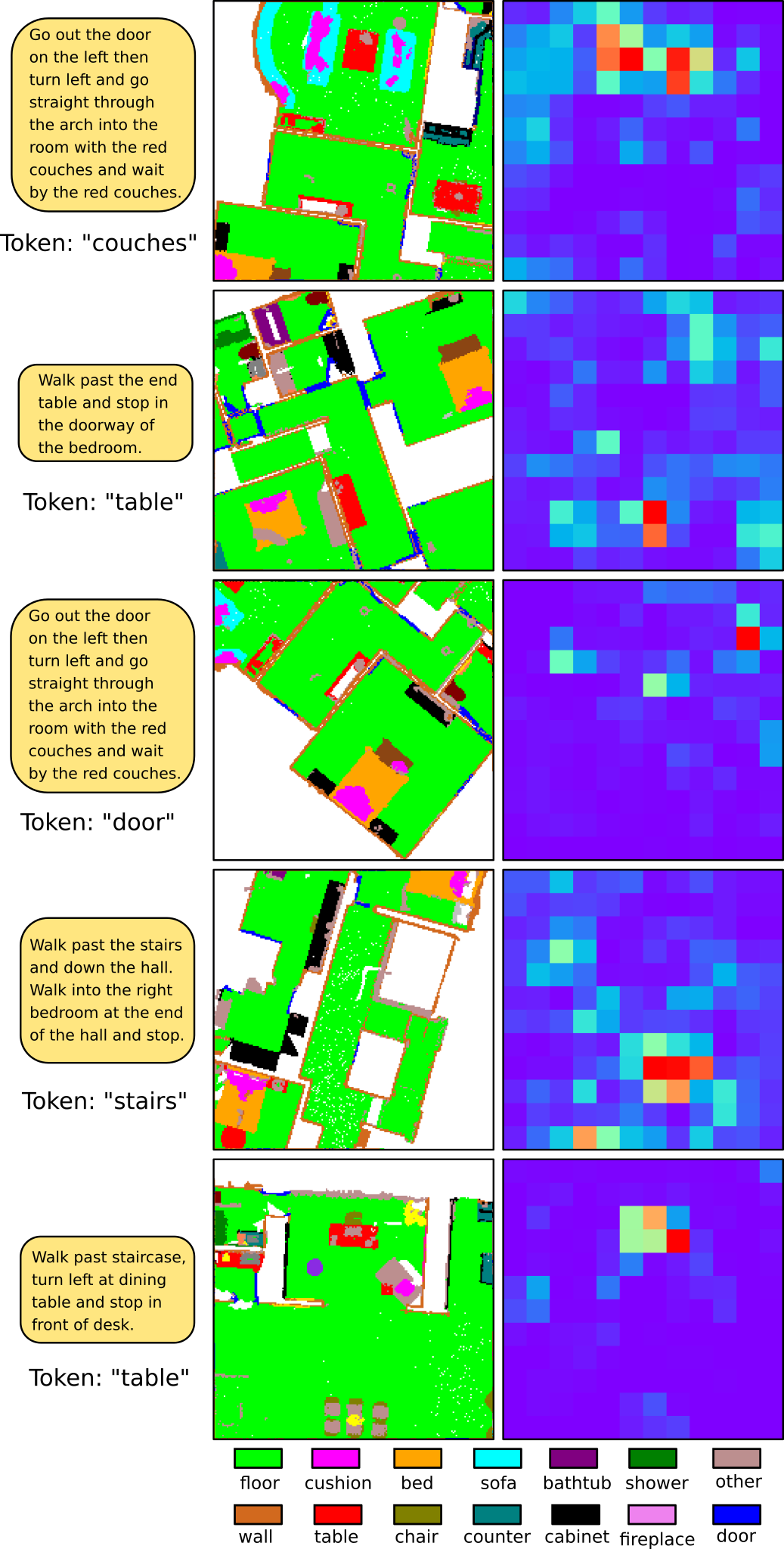}
    \caption{Visualization of the cross-modal attention representation between map and specific word tokens. The representation tends to focus on semantic areas of the map that correspond to the object referred to by the token. Note that in the example on the 4th row the representation focuses on the area where stairs are located, even though we do not use a specific semantic label for stairs in the map.}
    \label{fig:semantic_grounding_supp}
\end{figure}

\begin{figure*}[t]
    \centering
    \includegraphics[width=0.8\linewidth]{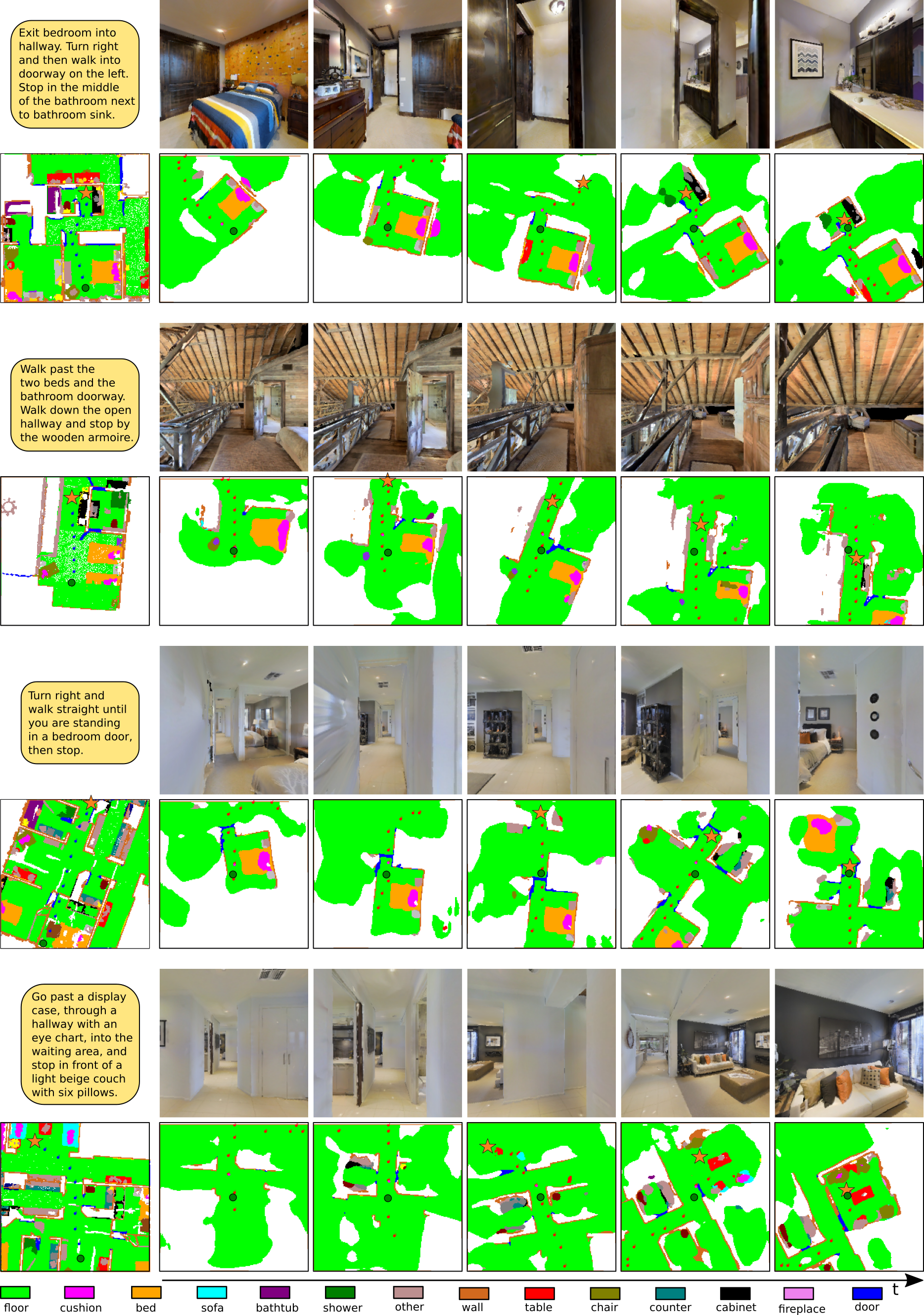}
    \caption{Navigation examples using our method \textit{CM\textsuperscript{2}} on \textit{val-seen} (first from top) and \textit{val-unseen} (last three). The top row of each example shows the RGB observations of the agent, while bottom shows the path prediction on the egocentric maps (the agent is in the middle looking upwards shown as the \textcolor{ao}{green circle}). The \textcolor{red}{red waypoints} represent our path prediction at the particular time-step. Observe that the goal, shown as an \textcolor{darkorange}{orange star}, is neither visible nor within the egocentric map at the beginning of the episodes. The ground-truth map and path are depicted in the bottom left corner.}
    \label{fig:nav_example_supp} 
\end{figure*}

Finally, we share additional visualizations of navigation episodes (Figure~\ref{fig:nav_example_supp}) and more examples of spatial and semantic grounding of the learned representations. Figure~\ref{fig:spatial_grounding_supp} shows the attention decoder output $H_t^s$ and Figure~\ref{fig:semantic_grounding_supp} presents more examples of the cross-modal attention.
See section 4.3 of the main paper for more details.


\end{document}